\documentclass[12pt,english]{article}
\usepackage[T1]{fontenc}
\usepackage[latin9]{inputenc}
\usepackage{array}
\usepackage{float}
\usepackage{multirow}
\usepackage{amsmath}
\usepackage{amssymb}
\usepackage{graphicx}

\makeatletter

\providecommand{\tabularnewline}{\\}

\newcommand{\lyxaddress}[1]{
\par {\raggedright #1
\vspace{1.4em}
\noindent\par}
}
\newenvironment{lyxlist}[1]
{\begin{list}{}
{\settowidth{\labelwidth}{#1}
 \setlength{\leftmargin}{\labelwidth}
 \addtolength{\leftmargin}{\labelsep}
 }}
{\end{list}}

\makeatother

\usepackage{babel}
\begin{document}

\title{Financial Risk and Returns Prediction with Modular Networked Learning}

\author{Carlos Pedro Gonçalves}
\maketitle

\lyxaddress{University of Lisbon, Instituto Superior de Ciências Sociais e Políticas
(ISCSP), cgoncalves@iscsp.ulisboa.pt}
\begin{abstract}
An artificial agent for financial risk and returns' prediction is
built with a modular cognitive system comprised of interconnected
recurrent neural networks, such that the agent learns to predict the
financial returns, and learns to predict the squared deviation around
these predicted returns. These two expectations are used to build
a volatility-sensitive interval prediction for financial returns,
which is evaluated on three major financial indices and shown to be
able to predict financial returns with higher than 80\% success rate
in interval prediction in both training and testing, raising into
question the Efficient Market Hypothesis. The agent is introduced
as an example of a class of artificial intelligent systems that are
equipped with a Modular Networked Learning cognitive system, defined
as an integrated networked system of machine learning modules, where
each module constitutes a functional unit that is trained for a given
specific task that solves a subproblem of a complex main problem expressed
as a network of linked subproblems. In the case of neural networks,
these systems function as a form of an ``artificial brain'', where
each module is like a specialized brain region comprised of a neural
network with a specific architecture.
\end{abstract}
\begin{lyxlist}{00.00.0000}
\item [{Keywords:}] Modular Networked Learning, Long Short-Term Memory,
Gated Recurrent Unit, Financial Turbulence, Financial Management
\end{lyxlist}

\section{Introduction}

In the context of artifical intelligence (AI), machine learning allows
for the development of intelligent systems that are able to learn
to solve problems \cite{key-1,key-2,key-3}. While a problem, where
an artificial agent is supplied with an information feature vector,
a target variable and a single optimization goal, can be solved by
programming the agent with some appropriate single machine learning
architecture, when dealing with more complex problems, that can be
expressed as a network of linked subproblems, each with its own target
and optimization goal, the artificial agent's cognitive architecture
needs to be expanded.

Using the framework of machine learning, this expansion becomes possible
by following the networked relations between the different subproblems.
In this case, the agent's cognitive structure can be designed in terms
of a Modular Networked Learning (MNL) system, where each module is
a cognitive unit that can be comprised of any machine learning architecture.
Each module can receive inputs from other modules to which it is linked
and can also receive external inputs.

Following the main problem's networked structure, each learning unit
is trained on a module-specific target, thus, instead of a single
regression or classification problem, an MNL system works with multiple
targets, which are linked to interconnected subproblems, functioning
as a modular interconnected system of learning units.

In this sense, an MNL system is an integrated system of machine learning
modules, where each learning module constitutes a functional unit
that is trained for a given specific task that solves a subproblem
of a complex main problem, comprised of different interconnected subproblems.
This creates a hierarchical modular cognitive system, where the modular
hierarchy comes out of the networked relations between the different
modules, that mirrors the networked relations between the different
subproblems.

In the case of artificial neural networks, for a general MNL system,
each module is a neural network and the MNL system is, itself, a neural
network, that is, a modular neural network where each module is a
neural network, thus, each learning module functions like a specialized
(artificial) ``brain region'' that is trained for a specific task
on a specific target, towards solving a particular subproblem. The
links between different neural networks, in the modular system, are,
as stated above, related to the networked structure of the main networked
problem that the MNL system is aimed at solving.

In finance, the application of MNL systems provides a different way
for financial modeling, management and analysis. Namely, traditional
financial theory uses two main statistical measures for portfolio
management and risk management: the expected returns and the variance
around the expected returns \cite{key-4,key-5,key-6}. However, the
main assumption of this theory is that financial markets are efficient,
which leads to the assumption that there is no way for which future
returns can be predicted based on past returns, with the last period's
price capitalized, using a fixed expected rate of return, being the
best predictor for price, this assumption led to the submartingale
statistical hypothesis for price movements \cite{key-7,key-8}.

Contrasting with this theory, financial management schools such as
fundamental analysis and technical analysis assumed that there is
information which can be used to anticipate price movements, and built
different trading rules based on past market data \cite{key-5,key-9,key-10}.

Applications of complex systems science and chaos theory to economics
and finance also raised the possibility of emergent nonlinear dynamics
coupled with stochastic noise to be present in financial returns'
dynamics, with consequences for financial theory development and financial
management \cite{key-7,key-8,key-11,key-12,key-13,key-14,key-15,key-16,key-17}.

Stochastic and deterministic volatility proposals \cite{key-7,key-8,key-14},
including fractal and multifractal risk and returns models \cite{key-18,key-19,key-20,key-16,key-17},
were introduced and shown to explain, at least to some extent, the
presence of emergent nonlinear dependence in financial turbulence,
affecting the variance around the expected financial returns.

In the 21st Century, with the rise of the FinTech industry and the
integration of machine learning in financial trading, leading to a
wave of automation in the financial industry \cite{key-22,key-23},
the issue of nonlinear market dynamics is a center stage problem for
both academia, financial community and regulators \cite{key-24,key-25,key-26}.

In the current work, we apply MNL to financial risk and returns prediction,
specifically, we address the two main parameters used as fixed by
the efficient market hypothesis proponents: the expected returns and
the variance around the expected returns. Without any prior knowledge
of the returns distribution, our goal is for an artificial agent to
be abe to:
\begin{enumerate}
\item Based upon a past time window of financial returns, to learn to produce
a conditional expectation for the next period's returns; 
\item Based upon the past track record of squared deviations of observed
versus predicted values, learn to produce a conditional expectation
for the next period's squared deviation around the next period's returns'
prediction, thus, learning to produce an expectation for its own prediction
error and, in this way, providing for a machine learning replacement
of a probabilistic time-varying variance;
\item Use the two expectations to produce a prediction interval for the
next period's returns that takes into account the predicted volatility
risk, measured by the square root of the next period's predicted squared
deviation.
\end{enumerate}
The third task is the main problem, which can be solved by using the
outputs of the first and second tasks (these are the two linked subproblems).
The first task can be solved by a module trained to predict the next
period's returns based on past returns, for a given time window, this
is the main prediction module. The second task, on the other hand,
has a hierarchical dependence upon the first task's module, since
it works with the notion of variance as an expected value for the
squared deviations around the expected value for a given target random
signal.

The evidence of presence of nonlinear volatility dynamics in financial
returns' dynamics \cite{key-7,key-8,key-14,key-15} means that the
conditional variance must change with time, the second task must,
therefore, employ a learning module that learns to predict the next
period's squared deviation around the first module's next period's
prediction, using a past record of squared deviations between observed
and predicted values.

Cognitively, this means that the artificial agent is capable of predicting
the target variable, in this case, the financial returns (the main
prediction module is specialized in this task), but the agent also
keeps an internal working memory record of past prediction performance,
in terms of the past squared deviations (this is a second module,
playing the role of a working memory unit), and uses this record to
predict the next period's squared deviation around the first module's
prediction (this is a third module and corresponds to a dispersion
prediction unit, which anticipates market volatility).

We, therefore, have an agent that learns not only to predict the target
variable but it also learns to anticipate a squared deviation for
its next period's prediction, quantifying an expected value for the
uncertainty associated with the temporal prediction task, which allows
the agent to emit a prediction interval centered on the first module's
prediction but that also takes into account the predicted deviation.
In this way, the agent is capable of outputing a volatility-adjusted
forward-looking time-varing prediction interval for the financial
returns.

While the dispersion prediction unit produces a machine learning version
of a conditional variance, for the interval prediction, we need to
take the square root of this module's output, thus, producing an interval,
around the expected returns, within a certain number of units of the
equivalent of a conditional standard deviation.

In terms of financial theory, we have a modular system that is capable
of capturing patterns in the returns' dynamics to produce an expectation
for the next period's returns' values, but it is also capable of using
the past track record of the squared deviations around its predictions
to anticipate the next period's market volatility, thus, incorporating
a second-order nonlinearity associated with market volatility. The
main prediction module is trained on the first task, while the dispersion
prediction module is trained on the second task.

The work is organized as follows, in section 2, we begin by addressing
modular hierarchical cognitive structures in neuroscience and in AI
research. Drawing on this research, in subsection 2.1, we introduce
the concept of an MNL system, addressing its general structure and
computational properties. In subsection 2.2, we address the financial
risk and returns' prediction agent, using the approach defined in
subsection 2.1. In subsection 2.3, we introduce two learning architectures
for the agent's modular units, one based on Long Short-Term Memory
(LSTM) networks, another based on Gated Recurrent Units (GRUs).

We show that both LSTMs and GRUs are, themselves, modular systems
of neural networks, so that, when we connect them in an MNL system
we get a more complex modular hierarchy, where each learning module
is itself a modular system.

When an MNL system is a hierarchical modular neural network, comprised
of learning modules that are, themselves, modular neural networks
trained on specific tasks, the cognitive system is not only able to
address the subproblem connectivity, associated with a complex problem,
but each subproblem is also adaptively addressed by a modular system
that is optimized for that specific subproblem. This generates a scaling
adaptivity from the local functional modular level up to the higher
scale of the whole artificial cognitive system.

In section 3, we compare these two architectures' performance on three
financial indices: S\&P 500, NASDAQ Composite and the Russell 2000.

In section 4, we conclude the work, discussing the implications of
the main results for finance.

\section{Modular Networked Learning for Financial Risk and Returns Prediction}

Modular networked cognition links adaptivity, functionality and information
flow towards the solution of complex problems. In a modular networked
cognitive system, the adaptive functionality of each module works
towards the overall system's adaptivity.

The information flow between the modules and the way in which the
modules are linked play a fundamental role in the whole system's adaptive
responses, since, considering any given module that depends upon other
modules for input, its adaptive responses, towards the fulfilling
of its functional role in the cognitive system, depend upon the adaptive
responses of those other modules, in this way, the evolutionary success
of the whole system depends upon the adaptive responses of each module.

In neuroscience, a modular structure in a brain can be identified
in terms of a localized topological structure that fulfills some function
in the organism's neural processing and cognition \cite{key-27,key-28,key-29}.
Modularity can be identified with brain regions or even multi-level
structures that are organized in terms of modular hierarchies \cite{key-27,key-28}.
In evolutionary terms, modularity, in biological brains, is argued
as promoting \emph{evolvability}, \emph{conserving wiring cost}, and
as a basis for s\emph{pecialized information processing and functionality}
\cite{key-29}.

In cognitive science, the argument for the importance of modularity
is not new, Fodor \cite{key-30}, for instance, proposed a modular
thesis of cognition. Changeux and Dehaene also expanded on cognitive
science's modular theses, from the perspective of evolutionary biology
and neuroscience \cite{key-31}. More recently, Kurzweil \cite{key-32}
built a pattern recognition approach to cognition, based upon hierarchical
modularity of pattern recognition modules, in order to approach a
computational model of the neocortex, aimed at research on AI. While
this later work addresses mostly one particular form of machine learning
architecture, in the current work, we propose the concept of an MNL
system as an integrated cognitive system comprised of modules that
can have any machine learning architecture. 

Our work draws a computational and cognitive basis from neuroscience
\cite{key-27,key-28,key-29,key-31}, in particular, on the importance
of a simultaneous functional specialization and functional interdependence,
within neurocognition, an interdependence that, as stated by Changeux
and Dehaene \cite{key-31}, holds for the different levels of neurofunctional
organization. Taking into account this framework, we now introduce
the concept of an MNL system.

\subsection{Modular Networked Learning Systems}

In its general form, an MNL system can be defined as an artificial
learning system comprised of a network of modules, where each learning
module is a cognitive unit that can have any machine learning architecture.
There are two general types of modules that we consider:
\begin{itemize}
\item Transformation units: whose task is to perform data transformation,
that can be used to extract learning features;
\item Learning units: that are aimed at learning to solve a specific task
and employ some machine learning architecture.
\end{itemize}
Each module can receive inputs from other modules to which it is linked
and can also receive external inputs. This leads to a second classification
for the modules:
\begin{itemize}
\item Sensory modules: whose input features are comprised only of external
data;
\item Internal modules: whose input features are comprised only of other
modules' outputs;
\item Mixed modules: whose input features are comprised of external data
and other modules' outputs.
\end{itemize}
Each learning unit, in an MNL system, is trained on a module-specific
target, producing a predicted value for that target, thus, instead
of a single regression or classification problem, an MNL system works
with multiple targets and (hierarchically) linked problems, functioning
as a hierarchical modular interconnected system.

The connections between the modules introduce a form of local hierarchical
information processing, where a given module may use other modules'
outputs as inputs for its respective functional effectiveness, which
is, in this case, the solution of its specific task. The whole system
thus addresses complex problems, involving multiple interconnected
subproblems.

The hierarchy comes out of the interconnections between the different
modules which, in turn, comes from the interconnection between the
different subproblems. Thus, rather than engineering the system by
taking the hierarchy as primitive, the modules (which, in the case
of a learning unit, is, itself, an artificial learning system) and
the connections between them are the primitive basis for the system's
engineering, the hierarchy comes out as a consequence of the connections
between the modules, which, in turn, reflects the nature of the complex
problems that the MNL system must solve.

The training protocols for the MNL system must follow the inter-module
linkages, therefore, the sensory modules must be trained first, since
they do not receive any internal input, only working on external data,
then, the training must follow the sequence of linkages, so that,
at each training stage, a specific learning unit is only trained after
all its input learning units' training, since it must receive the
outputs of these units, in order to be able to learn, and it must
learn to respond to already optimized input learning units.

Indeed, in the present work, we are dealing with an example of cooperative
networks that are trained to solve different interconnected objectives,
building on each other's work. In this context, in order to be properly
optimized, a given learning unit needs the learning units, upon which
it depends for its functionality, to already be trained.

Since any machine learning architecture can be chosen for the modules,
depending on the architectures used, we can have a different modular
system. An MNL system comprised only of neural networks has, however,
a specific property: the system is, itself, a neural network, that
is, we have a modular system of neural networks that is, itself, a
neural network, which adds another level to deep learning that can
be characterized as form of deep modular hierarchical learning, since,
each neural network is specialized in a subproblem, but the networks'
processing is linked, which means that, at the higher levels (higher
cognitive functions), the modules need to work with the outputs of
the processing of other modules, beginning with the sensory modules.
This allows the development of \emph{deep modular networked learning},
where each learning unit can use a (possibly different) deep learning
architecture.

In the case of recurrent neural networks, whose learning units are
recurrent neural networks, the connection between two learning units
requires an intermediate working memory unit, which can be defined
as a transformation unit. We will show an example of this in the next
section.

Any type of neural network can be used for different modules, the
network architecture for each module depends upon the specific problem
that it is aimed at solving. The same holds for other machine learning
architectures.

To exemplify the application of MNL systems to interconnected problems,
we will now apply MNL to financial risk and returns prediction.

\subsection{A Financial Risk and Returns' Predictor Agent Using Modular Networked
Learning}

The efficient market hypothesis (EMH), introduced in 1970 by Fama
\cite{key-34}, states that financial assets' prices always fully
reflect all available information, this hypothesis was integrated
in modern financial theory, which developed from the 1950s to the
1980s and 1990s \cite{key-4,key-5,key-6,key-19,key-35}. There are
three versions of the EMH, as addressed by Fama \cite{key-34}: the
weak version assumes that all historical price information is already
reflected in prices, the semi-strong version extends the hypothesis
to include all the publicly available information and the strong version
extends the hypothesis to include all relevant information (public
and private), which means, in this last case, that one cannot consistently
profit from insider trading.

The EMH versions are nested, in the sense that the if the semi-strong
version fails the strong version fails, even if the weak version may
not fail, on the other hand, if the weak version fails, the strong
and semi-strong versions also fail, and the EMH itself fails.

The weak version's main statistical implication is that we cannot
use past price information to predict the next period's price. This
implication actually comes out of Fama's argument regarding the empirical
testability of the EMH, which led the author to recover the concept
of fair game, and propose the martingale and submartingale models
as a working basis for statistically testing the EMH \cite{key-34}. 

In the statistical assumption that we cannot use past price information
to predict the next period's price, the EMH is in opposition to other
schools of financial management, in particular, to technical analysis,
which assumes that there is relevant information in past price trends
that can help predict the next period's price \cite{key-9,key-19,key-35}.

If the EMH is correct, then, that means that we cannot predict the
next period's price based on past trends, namely, statistically, for
a financial price series $S_{t}$, given any past price time window
$\left\{ S_{t},S_{t-1},S_{t-2},...,S_{t-\tau}\right\} $, the expected
value of the price at time $t+1$ obeys the submartingale condition
\cite{key-6,key-34}:
\begin{equation}
E\left[S_{t+1}|S_{t},S_{t-1},...,S_{t-\tau}\right]=S_{t}e^{\mu}
\end{equation}
that is, the best prediction for the next period's price is the current
price capitalized by one period, using a fixed exponential rate of
return $\mu\geq0$.

If we consider the logarithmic returns, connecting the price at time
$t+1$ to the price at time $t$, we get: 
\begin{equation}
S_{t+1}=S_{t}e^{r_{t+1}}
\end{equation}
where $r_{t+1}$ is a random variable, then, the submartingale condition
implies that:
\begin{equation}
E\left[r_{t+1}|r_{t},r_{t-1},...,r_{t-\tau}\right]=\mu
\end{equation}
Indeed, if the conditional expected value of the returns over a past
time window was not equal to a fixed value $\mu$, thus, independent
from past returns, then, we could use the price value at $t-\tau$,
$S_{t-\tau}$, and the sequence $r_{t},r_{t-1},...,r_{t-\tau}$ to
anticipate the next period's financial price.

A second early assumption used in modern financial theory, in particular,
in modern portfolio theory and option pricing theory \cite{key-4,key-5,key-6,key-19},
is that the logarithmic returns series have a fixed variance, so that
we also get the condition:
\begin{equation}
E\left[\left(r_{t+1}-\mu\right)^{2}|r_{t},r_{t-1},...,r_{t-\tau}\right]=\sigma^{2}
\end{equation}
With the development of chaos theory and the complexity approach to
economics and finance \cite{key-11,key-12,key-13,key-14,key-15,key-16},
mainly, in the 1980s and 1990s, the possibility of emergent complex
nonlinear dynamics, linking trading strategies and multiagent adaptation,
led to other hypotheses and market theories, namely, the coherent
market hypothesis (CMH), introduced by Vaga \cite{key-36}, which
assumes that markets undergo phase transitions and addresses these
phase transitions applying statistical mechanics, and the adaptive
market hypothesis (AMH), introduced by Lo \cite{key-37}, who applied
evolutionary biology, complex systems science and neuroscience, raising
the argument for the connection between the traders' adaptive dynamics
and emergent market patterns. Both hypotheses are convergent on one
point: the possibility of emergent complex nonlinear dynamics.

In particular, emergent complex nonlinear dynamics can take place,
leading to both the expected value and variance changing with time
as a basic consequence of investors' adaptive responses to risk and
returns, so that we get an emergent functional dependence:
\begin{equation}
E\left[r_{t+1}|r_{t},r_{t-1},...,r_{t-\tau}\right]=\mu_{t+1}=f_{0}\left(r_{t},r_{t-1},...,r_{t-\tau}\right)
\end{equation}
\begin{equation}
E\left[\left(r_{t+1}-\mu_{t+1}\right)^{2}|r_{t},r_{t-1},...,r_{t-\tau}\right]=\sigma_{t+1}^{2}=f_{1}\left(r_{t},r_{t-1},...,r_{t-\tau}\right)
\end{equation}
These functions are, however, unknown, with different competing models
being able to produce similar turbulence patterns \cite{key-15,key-16,key-17,key-18,key-19,key-20,key-38}.

When there is turbulence and strong nonlinearities, the conditional
probabilities can change from time period to time period in unstable
ways, a major argument raised by Vaga \cite{key-36}. In this case,
without the knowledge of the underlying probability structure and
dynamical rules, the human analyst is faced with different possible
alternatives that might have generated the same data structure, thus,
an underlying problem for financial management arises due to the complex
nonlinear dynamics present in the data, since this nonlinear dynamics
can have a complex functional form and/or be influenced by stochastic
factors and, also, it affects two key parameters: how much one can
expect to get as returns on an investment, and the level of volatility
risk associated with that investment.

Key methodologies used for financial management, such as equivalent
martingale measures applied to financial derivatives' risk-neutral
valuation, can be invalidated by changes in the market volatility,
linked to a nonlinear evolution of the variance, which can lead to
multiple equivalent measures \cite{key-6}.

Whenever there are strong dynamical instabilities, it becomes difficult
to estimate a probability distribution by applying a frequencist approach,
so that we are, in fact, dealing with a decision problem under uncertainty,
where the financial manager needs to form an expectation for the financial
returns and an evaluation of the returns' variance, producing a calculatory
basis for the expected value and variance for time $t+1$, given the
information available at time $t$, but has no access to the underlying
probability nor to the dynamical rules.

An artificial agent built with an MNL system can be applied to these
contexts, since it is effective in capturing complex nonlinear dynamical
relations between predictors and target variables.

In this way, the probability-based expected value for the next period's
returns (equation (5)) is replaced, in the context of uncertainty,
by the machine learning-based expected value for the next period's
returns, produced by a module that is specialized in predicting the
the financial returns' next period's value. We call this module the
main prediction unit (MPU), it is a sensory module, since it works
with the external data signal, and a learning unit aimed at learning
to predict the financial returns' next period's value.

Likewise, the variance, which is the expected value for the squared
deviation around the expected value (equation (6)), is replaced, in
the context of uncertainty, by the machine learning-based expected
value for the next period's squared deviation around the machine learning-based
expected returns, which implies the need for another learning unit
that is trained on that specific task, we call this module the dispersion
prediction unit (DPU).

The two modules are linked in a modular network. This network must
actually be comprised of three modules: two learning units (the MPU
and the DPU) plus one additional sliding window memory unit which
is a transformation unit, keeping track of the past record of squared
deviations. The first unit is aimed at producing an expectation for
the financial returns $r_{t+1}$, given the last $\tau+1$ observations,
so that we replace the expected value in equation (5) by the output
of the learning unit: 
\begin{equation}
\mu_{t+1}=\Phi_{MPU}\left(\mathbf{r}_{t}\right)
\end{equation}
where $\mathbf{r}_{t}\in\mathbb{R}^{\tau+1}$ is a sliding window
vector of the form: 

\begin{equation}
\mathbf{r}_{t}=\left(\begin{array}{c}
r_{t-\tau}\\
\vdots\\
r_{t-1}\\
r_{t}
\end{array}\right)
\end{equation}
and $\Phi_{MPU}$ is the nonlinear mapping produced by a neural network
linking the past returns information, expressed in the vector $\mathbf{r}_{t}$,
to the expected value for the next period's returns $\mu_{t+1}$,
approximating the unknown function $f_{0}$ in equation (5).

Now, since the input $\mathbf{r}_{t}$ has an internal sequential
structure, as a sliding window vector signal, this means that, by
using a recurrent neural network for the MPU, the processing of the
input will take into account the sequential structure present in the
input to the MPU. This is relevant because of the evidence that financial
returns go through regime changes, a point raised in complex systems
science approaches to finance \cite{key-13,key-14,key-15,key-16,key-17,key-18,key-36},
which means that the sequence of values in $\mathbf{r}_{t}$ matters,
recurrent neural networks (RNNs) capture this type of sequential dependence
in data. The MPU may, thus, predict positive, negative or null returns,
which means that the market price can be expected to rise, remain
unchanged or fall, depending on the sequential patterns exhibited
by the past returns. This is in stark contrast to the sub-martingale
property assumed by the EMH.

Let, then, for a decreasing sequential index\footnote{By considering a decreasing index sequence we are moving forward in
time from the earliest element in the sequence to the latest.} $i=\tau,...,1,0$, $\hat{\Pi}_{i}$ be the projection operator on
the index $i$ component of a $\tau+1$ size vector, so that for the
input $\mathbf{r}_{t}$, we have:
\begin{equation}
\hat{\Pi}_{i}\mathbf{r}_{t}=r_{t-i}
\end{equation}
and let $\mathbf{h}_{1,t}\in\mathbb{R}^{d_{1}}$ be the $d_{1}$-dimensional
hidden state of the RNN for the first module (the MPU) at time $t$,
where we use the lower index $1$ as the module's index. For a time
window of $\tau+1$ size, the RNN generates a sequence of hidden states,
so that we have the mapping of the sequence $\left\{ \hat{\Pi}_{i}\mathbf{r}_{t}\right\} _{i=\tau}^{0}$
to a corresponding sequence of hidden states $\left\{ \mathbf{h}_{1,t-i}\right\} _{i=\tau}^{0}$,
where $\mathbf{h}_{1,t-i}$ is updated in accordance to some pointwise
nonlinear activation function $\Phi_{1}$ that results from the computation
of the RNN, namely, we have the general update rule:
\begin{equation}
\mathbf{h}_{1,t-i}=\Phi_{1}\left(\mathbf{h}_{1,t-i-1},\hat{\Pi}_{i}\mathbf{r}_{t}\right)
\end{equation}
that is, the network sequentially projects each component of the input
data vector and updates the hidden state, depending upon its previous
hidden state and the projected value of the input vector.

Now, given the sequence of hidden states, the last element of the
sequence is fed forward to a single output neuron, so that the final
prediction is obtained by the feedforward rule: 
\begin{equation}
\mu_{t+1}=\Theta_{1}\left(\mathbf{W}_{1}^{T}\mathbf{h}_{1,t}+b_{1}\right)
\end{equation}
where $\mathbf{W}_{1}\in\mathbb{R}^{d_{1}}$ is a synaptic weights
vector, $b_{1}\in\mathbb{R}$ is a bias and $\Theta_{1}$ is a nonlinear
activation function for the final feedforward connection. In this
way, the MPU is trained on a regression problem, such that the unknown
function in equation (5) is replaced by the module's predictions.
Formally, using the general scheme in equations (10) and (11), we
get the nested structure:
\begin{equation}
\begin{aligned}\mu_{t+1}=\Phi_{MPU}\left(\mathbf{r}_{t}\right)=\\
\Theta_{1}\left[\mathbf{W}_{1}^{T}\Phi_{1}\left(\left(...\Phi_{1}\left(\Phi_{1}\left(\mathbf{h}_{1,t-\tau-1},\hat{\Pi}_{\tau}\mathbf{r}_{t}\right),\hat{\Pi}_{\tau-1}\mathbf{r}_{t}\right)\right),\hat{\Pi}_{0}\mathbf{r}_{t}\right)+b_{1}\right]
\end{aligned}
\end{equation}

The RNN scheme incorporates a basic financial adaptation dynamics,
where the patterns in the past returns' sequential values affect the
trading patterns and, therefore, the next returns' value. This is
a basic feedback dynamics in the emergent market cognition and activity.
The artificial agent incorporates this in its processing of its expected
value for the next period's returns.

Now, since financial time series exhibit evidence of time-varying
volatility in the returns' series, with long memory and clustering
of turbulence \cite{key-13,key-14,key-15,key-16,key-17,key-18,key-19},
the deviations around expected returns must be time-varying and patterns
may be present in the squared deviations around an expected value,
so that we do not get a stationary variance \cite{key-13,key-14,key-15,key-18,key-19,key-20}.

Financially, this means that while market agents trade based on expectations
regarding returns, they also incorporate market volatility risk in
trading decisions, which introduces a feedback with regards to returns
and volatility, linked to the risk of fluctuations around the expected
value. This also means that the market will be sensitive to changes
in volatility and these changes end up influencing trading decisions
and, thus, the volatility dynamics itself.

Therefore, a second level of expectations needs to be computed by
the artificial agent, in order to reflect a basic market computation
of patterns in volatility, that is, in order to capture emergent dynamical
patterns in volatility, the artificial agent needs to have a cognitive
system that is capable of producing an expectation for volatility,
which we will measure in terms of an expected squared deviation around
the next period's agent's prediction.

In this way, the agent must be capable of incorporating past volatility
data into a prediction of the next period's squared deviation around
its own expectation, such that the artificial agent is capable of
producing a machine learning equivalent of a forward looking variance,
replacing equation (6) with a machine learning-based variance. In
order to do this, we need the other two modules, the working memory
module and the DPU.

The working memory module is a transformation unit, in the form of
a sliding window working memory neural network, that keeps track of
the past prediction performance in terms of squared deviations. This
network has $\tau+1$ neurons linked in chain. The state of the network
at time $t$ is described by a vector $\mathbf{s}_{t}^{2}\in\mathbb{R}^{\tau+1}$
defined as:
\begin{equation}
\mathbf{s}_{t}^{2}=\left(\begin{array}{c}
\left(r_{t-\tau}-\mu_{t-\tau}\right)^{2}\\
\vdots\\
\left(r_{t-1}-\mu_{t-1}\right)^{2}\\
\left(r_{t}-\mu_{t}\right)^{2}
\end{array}\right)
\end{equation}
Now using the projection operators, so that $\hat{\Pi}_{i}\mathbf{s}_{t}^{2}=\left(r_{t-i}-\mu_{t-i}\right)^{2}$,
the state transition from $\mathbf{s}_{t}^{2}$ to $\mathbf{s}_{t+1}^{2}$
is, locally, described as: 
\begin{equation}
\hat{\Pi}_{i}\mathbf{s}_{t+1}^{2}=\begin{cases}
\hat{\Pi}_{i-1}\mathbf{s}_{t}^{2}, & i\neq0\\
\left(\hat{\Pi}_{0}\mathbf{r}_{t+1}-\mu_{t+1}\right)^{2}, & i=0
\end{cases}
\end{equation}
for $i=0,1,2,...,\tau$, which leads to the sliding window transition:
\begin{equation}
\left(\begin{array}{c}
\left(r_{t-\tau}-\mu_{t-\tau}\right)^{2}\\
\vdots\\
\left(r_{t-1}-\mu_{t-1}\right)^{2}\\
\left(r_{t}-\mu_{t}\right)^{2}
\end{array}\right)\rightarrow\left(\begin{array}{c}
\left(r_{t+1-\tau}-\mu_{t+1-\tau}\right)^{2}\\
\vdots\\
\left(r_{t}-\mu_{t}\right)^{2}\\
\left(r_{t+1}-\mu_{t+1}\right)^{2}
\end{array}\right)
\end{equation}

Therefore, the last neuron in the chain records the last squared deviation,
and the remaining neurons implement a sliding window operation, leading
to a mixed module, which always keeps track of the past prediction
performance, in terms of squared deviations, over the time window.
The agent is therefore, keeping track of its own past squared prediction
errors.

The third module, is an internal module, that uses the input from
the sliding window working memory neural network in order to predict
the next period's squared deviation. Thus, given a training sequence
of pairs of the form $\left(\mathbf{s}_{t}^{2},s_{t+1}^{2}\right)$,
with $s_{t+1}^{2}=\left(r_{t+1}-\mu_{t+1}\right)^{2}$, the third
module learns to predict the next period's dispersion around the expected
value produced by the first module, which justifies the name dispersion
prediction unit (DPU).

Since the sequential patterns of volatility matter for the task of
predicting volatility, in particular, changes in volatility and the
clustering of turbulence \cite{key-14,key-15,key-18,key-19} need
to be processed by the agent, then, as in the case of the MPU, we
also need to work with a recurrent neural network, therefore, as in
the case of the MPU, we also get the sequentially nested structure:
\begin{equation}
\begin{aligned}\sigma_{t+1}^{2}=\Phi_{DPU}\left(\mathbf{s}_{t}^{2}\right)=\\
\Theta_{3}\left[\mathbf{W}_{3}^{T}\Phi_{3}\left(\left(...\Phi_{3}\left(\Phi_{3}\left(\mathbf{h}_{3,t-\tau-1},\hat{\Pi}_{\tau}\mathbf{s}_{t}^{2}\right),\hat{\Pi}_{\tau-1}\mathbf{s}_{t}^{2}\right)\right),\hat{\Pi}_{0}\mathbf{s}_{t}^{2}\right)+b_{3}\right]
\end{aligned}
\end{equation}
where $\mathbf{W}_{3}\in\mathbb{R}^{d_{3}}$ is a synaptic weights
vector, $b_{3}\in\mathbb{R}$ is a bias and $\Theta_{3}$ is a pointwise
nonlinear activation function for the final feedforward connection.
The training of the third unit replaces the unknown function in equation
(6) by the learning unit's predictions.

Regarding the last nonlinear activation, for the MPU, we will set
$\Theta_{1}$, in equation (12), to the $\tanh$ activation, since
its prediction target (the financial returns) can be both positive
and negative, financially, also, the $\tanh$ structure may be useful
to capture market correction movements. In the case of the DPU, the
target cannot be negative, so we will set $\Theta_{3}$ to a ReLU
activation\footnote{Another alternative would be a sigmoid activation, however, the sigmoid
dampens for higher values of the input, while the ReLU obeys a linear
rule for a positive input, which is better for capturing jumps in
volatility. The squared deviation along with the last ReLU layer allow
one to incorporate, in a single dynamical recurrent model, the scaling
of nonlinearities \cite{key-18,key-19} and generalized auto-regressive
conditional heteroskedasticity (GARCH)-type squared dynamical dependence
patterns \cite{key-13,key-14,key-15}.}. 

A relevant point, characteristic of a modular system, is that there
is a scaling of nonlinearities in equation (16), indeed, since the
squared errors, received from the sliding window (working) memory
block, incorporate the past outputs of the first module, we have a
hierarchical scaling of nonlinear transformations of the original
returns signal. This is a useful point, because it allows the MNL
system to capture patterns associated with strong nonlinearities in
the original signal, in order to synthesize an expected returns value
and a volatility risk expectation.

The agent's MNL system is, thus, not only capable of producing a prediction
for a target, but it is also able to address its own past prediction
performance and to anticipate by how much it may fail in its next
step prediction. This is the result of the linkage between the three
modules, where the first module provides the prediction for the target,
the second module produces a working memory record of the past squared
deviations and the third module uses that record to anticipate the
next step squared deviation, effectively incorporating a concept of
expected squared dispersion around an expected value (the concept
of variance). Using this expected squared dispersion, a machine learning-based
standard deviation can be extracted from equation (16) by taking the
square root:
\begin{equation}
\sigma_{t+1}=\sqrt{\sigma_{t+1}^{2}}=\sqrt{\Phi_{DPU}\left(\mathbf{s}_{t}^{2}\right)}
\end{equation}

A risk and returns' prediction agent, defined as an artificial agent
with the above built-in MNL system can, thus, reduce the uncertainty
associated with the functional dynamical relations in equations (5)
and (6), and provide a calculatory basis for expected returns and
risk assessment.

In particular, in what regards the prediction of the next period's
financial returns, the agent, at time $t$, can use the outputs of
the MPU and DPU to produce a prediction interval for time $t+1$,
that takes into account both the returns' expected value as well as
the time-varying machine learning-based standard deviation defined
as the square root of the DPU's output, so that, instead of a single
point prediction, the agent provides an interval prediction that also
incorporates volatility risk prediction, formally:

\begin{equation}
I_{l,t+1}=\left]\mu_{t+1}-l\sigma_{t+1};\mu_{t+1}+l\sigma_{t+1}\right[
\end{equation}
The agent is, thus, reporting a prediction within $l$ units of $\sigma_{t+1}$.
The output of the MPU sets the center for the interval, while the
output of the DPU sets the scale for the reported uncertainty in the
agent's prediction and, thus, for the predicted market volatility.
While the uncertainty level $l$ is fixed, $\sigma_{t+1}$ is an anticipated
uncertainty given a time window of data, which means that the interval's
amplitude adapts to the market data.

In terms of interval prediction performance, given a sample sequence
of length $T$, the proportion of cases that fall within the interval
bounds can be calculated as follows: 
\begin{equation}
p_{l}=\frac{\#\left\{ t:r_{t}\in I_{l,t},t=t_{1},t_{2},...,T\right\} }{T}
\end{equation}

The success in prediction depends upon the agent's MNL system's success
in expectation formation, which, in turn, depends upon the ability
of the agent's MNL system to capture main nonlinear patterns in the
returns and squared deviations' dynamics, producing expectations based
upon these patterns. We will compare, in the next section, the performance
of two main recurrent architectures LSTM and GRU, which we now address.

\subsection{Learning Units' Architectures}

The architectures LSTM \cite{key-39,key-40} and GRU \cite{key-41,key-42}
are two recurrent neural network architectures that are, themselves,
modular networks comprised of interacting neural networks. When the
RNNs assigned to the MPU and DPU are LSTM or GRU neural networks,
we get a modular system whose learning units are, thus, also, modular
systems, so that a linked scaling modularity is obtained from the
individual learning units up to the entire MNL system.

While, at the level of the individual learning units, the modularity,
in this case, is aimed at the adaptive memory management that captures
main temporal patterns in the input data and is trained at predicting
a single target, at the level of the entire modular networked cognitive
system the agent's learning units are each trained on a different
target. This distinction between target specific modularity and multitarget
modularity is important when dealing with an MNL system comprised
of different target-specific modular neural networks.

The artificial agent's cognitive system, thus, becomes an integrated
neural network with a scaling modular organization where each specialized
learning unit is itself a modular neural network. To place this point
more formally, let us consider the main equations for the LSTM.

\subsubsection{Long Short-Term Memory Architecture}

Considering an LSTM architecture, the hidden state update rule depends
upon a memory block, comprised of a number of $d_{k}$ memory cells,
where, as previously, we set the module index $k=1$ for the MPU and
$k=3$ for the DPU, formally, for a decreasing index $i=\tau,...,1,0$
the update rule depends upon output gate neurons and the memory cells
as follows:
\begin{equation}
\mathbf{h}_{k,t-i}=\mathbf{o}_{k,t-i}\odot\tanh\left(\mathbf{c}_{k,t-i}\right)
\end{equation}
where $\odot$ stands for the pointwise multiplication and $\tanh$
is the pointwise hyperbolic tangent activation function, $\mathbf{c}_{k,t-i}\in\mathbb{R}^{d_{k}}$
is the vector for the LSTM memory cells' state at time $t-i$ and
$\mathbf{o}_{k,t-i}\in\mathbb{R}^{d_{k}}$ is an output gate vector
that modulates the hidden layer's response to the neural activation
resulting from the memory cells' signal. Both the output gate and
the memory cells will depend upon the input data and the previous
hidden layer's state.

The output gate is an adaptive neural response modulator, comprised
of $d_{k}$ neurons, that modulates (pointwise) multiplicatively the
response of the hidden layer neurons to the LSTM memory cells, thus,
the neurons that comprise this output gate play the role of what can
be considered memory response modulator neurons. These neurons also
form the output layer of a neural network that receives inputs from:
external input data, the previous hidden state and the memory cells'
state (peephole connection) plus a bias, so that we have the following
update rules for the MPU and DPU's memory response modulator neurons,
respectively:

\begin{equation}
\mathbf{o}_{1,t-i}=\textrm{sigm}\left(\left(\hat{\Pi}_{i}\mathbf{r}_{t}\right)\mathbf{W}_{1,o}+\mathbf{R}_{1,o}\mathbf{h}_{1,t-i-1}+\mathbf{p}_{1,o}\odot\mathbf{c}_{1,t-i}+\mathbf{b}_{1,o}\right)
\end{equation}
\begin{equation}
\mathbf{o}_{3,t-i}=\textrm{sigm}\left(\left(\hat{\Pi}_{i}\mathbf{s}_{t}^{2}\right)\mathbf{W}_{3,o}+\mathbf{R}_{3,o}\mathbf{h}_{3,t-i-1}+\mathbf{p}_{3,o}\odot\mathbf{c}_{3,t-i}+\mathbf{b}_{3,o}\right)
\end{equation}
where $\textrm{sigm}$ is the sigmoid activation function\footnote{We use this notation since the symbol $\sigma$ is already being used
for the standard deviation.}.

At each update, the local network, supporting the memory response
modulator neurons' computation for the MPU, processes the past financial
returns at step $t-i$, with associated synaptic weights vector $\mathbf{W}_{1,o}\in\mathbb{R}^{d_{1}}$,
which explains the first term in equation (21). Formally, we have:

\begin{equation}
\left(\hat{\Pi}_{i}\mathbf{r}_{t}\right)\mathbf{W}_{1,o}=r_{t-i}\mathbf{W}_{1,o}=\left(\begin{array}{c}
r_{t-i}w_{1,o}^{1}\\
r_{t-i}w_{1,o}^{2}\\
\vdots\\
r_{t-i}w_{1,o}^{d_{1}}
\end{array}\right)
\end{equation}
The first term for the DPU, on the other hand, processes the past
squared deviations at step $t-i$, stored in the sliding window memory
block, with associated synaptic weights vector $\mathbf{W}_{3,o}\in\mathbb{R}^{d_{3}}$,
so that we get, for the first term in equation (22), the following
result: 
\begin{equation}
\left(\hat{\Pi}_{i}\mathbf{s}_{t}^{2}\right)\mathbf{W}_{3,o}=\left(r_{t-i}-\mu_{t-i}\right)^{2}\mathbf{W}_{3,o}=\left(\begin{array}{c}
\left(r_{t-i}-\mu_{t-i}\right)^{2}w_{3,o}^{1}\\
\left(r_{t-i}-\mu_{t-i}\right)^{2}w_{3,o}^{2}\\
\vdots\\
\left(r_{t-i}-\mu_{t-i}\right)^{2}w_{3,o}^{d_{3}}
\end{array}\right)
\end{equation}

The remaining structure of the two modules is the same: each module's
memory response modulator neurons receive a recurrent input from the
previous hidden state $\mathbf{h}_{k,t-i-1}$, with a corresponding
synaptic weights matrix $\mathbf{R}_{k,o}\in\mathbb{R}^{d_{k}\times d_{k}}$.
Assuming a peephole connection, the memory response modulator neurons
also receive an input from the memory cells with peephole weights
$\mathbf{p}_{k,o}\in\mathbb{R}^{d_{k}}$. The pointwise multiplication
means that the relation for peephole connections is one-to-one. Finally,
both modules add a bias $\mathbf{b}_{k,o}\in\mathbb{R}^{d_{k}}$,
before applying the pointwise sigmoid activation function.

This is the main structure for the output gate. Now, for each learning
unit, the corresponding memory cells update their state in accordance
with the following rule: 
\begin{equation}
\mathbf{c}_{k,t-i}=\mathbf{i}_{k,t-i}\odot\mathbf{z}_{k,t-i}+\mathbf{f}_{k,t-i}\odot\mathbf{c}_{k,t-i-1}
\end{equation}

In this case, $\mathbf{z}_{k,t-i}\in\mathbb{R}^{d_{k}}$ is an input
layer to the cells that makes the memory cells update their state,
processing a new memory content, the impact of this input layer is
modulated by the input gate, $\mathbf{i}_{k,t-i}\in\mathbb{R}^{d_{k}}$,
which determines the degree to which the cells update their state,
incorporating the new memory content in their processing.

Likewise, there is a tendency of the memory cells not to process new
memories, like a short-term memory forgetfulness which is incorporated
in the second pointwise product that comprises the memory cells' update
rule, which is comprised of the previous step memory cells' state
$\mathbf{c}_{k,t-i-1}$, pointwise multiplied by the forget gate $\mathbf{f}_{k,t-i}\in\mathbb{R}^{d_{k}}$,
which modulates the degree of response of the memory cells to their
previous state. In the limit case, if $\mathbf{i}_{k,t-i}=\mathbf{0}_{d_{k}}$
and $\mathbf{f}_{k,t-i}=\mathbf{1}_{d_{k}}$, where $\mathbf{0}_{d_{k}}$
is a $d_{k}$-sized column vector of zeroes and $\mathbf{1}_{d_{k}}$
is a $d_{k}$-sized column vector of ones, the memory cells' state
would remain unchanged $\mathbf{c}_{k,t-i}=\mathbf{c}_{k,t-i-1}$,
which means that the network would be unable to incorporate new memory
contents in its dynamical memory processing.

For the LSTM architecture, the new memory contents plus the input
and forget gates are, each, the output layer of a respective neural
network. The input and forget gates follow the same rules as the output
gate $\mathbf{o}_{k,t-i}$, with the exception that the peephole connection
depends upon the previous rather than the updated state of the memory
cells, thus, for the MPU, we have: 
\begin{equation}
\mathbf{i}_{1,t-i}=\textrm{sigm}\left(\left(\hat{\Pi}_{i}\mathbf{r}_{t}\right)\mathbf{W}_{1,in}+\mathbf{R}_{1,in}\mathbf{h}_{1,t-i-1}+\mathbf{p}_{1,in}\odot\mathbf{c}_{1,t-i-1}+\mathbf{b}_{1,in}\right)
\end{equation}
\begin{equation}
\mathbf{f}_{1,t-i}=\textrm{sigm}\left(\left(\hat{\Pi}_{i}\mathbf{r}_{t}\right)\mathbf{W}_{1,f}+\mathbf{R}_{1,f}\mathbf{h}_{1,t-i-1}+\mathbf{p}_{1,f}\odot\mathbf{c}_{1,t-i-1}+\mathbf{b}_{1,f}\right)
\end{equation}
while, for the DPU, we have:
\begin{equation}
\mathbf{i}_{3,t-i}=\textrm{sigm}\left(\left(\hat{\Pi}_{i}\mathbf{s}_{t}^{2}\right)\mathbf{W}_{3,in}+\mathbf{R}_{3,in}\mathbf{h}_{3,t-i-1}+\mathbf{p}_{3,in}\odot\mathbf{c}_{3,t-i-1}+\mathbf{b}_{3,in}\right)
\end{equation}
\begin{equation}
\mathbf{f}_{3,t-i}=\textrm{sigm}\left(\left(\hat{\Pi}_{i}\mathbf{s}_{t}^{2}\right)\mathbf{W}_{3,f}+\mathbf{R}_{3,f}\mathbf{h}_{3,t-i-1}+\mathbf{p}_{3,f}\odot\mathbf{c}_{3,t-i-1}+\mathbf{b}_{3,f}\right)
\end{equation}

The new memory contents for the two modules' memory cells are, respectively,
udpated in accordance with the rules:

\begin{equation}
\mathbf{z}_{1,t-i}=\tanh\left(\left(\hat{\Pi}_{i}\mathbf{r}_{t}\right)\mathbf{W}_{1,z}+\mathbf{R}_{1,z}\mathbf{h}_{1,t-i-1}+\mathbf{b}_{1,z}\right)
\end{equation}
\begin{equation}
\mathbf{z}_{3,t-i}=\tanh\left(\left(\hat{\Pi}_{i}\mathbf{s}_{t}^{2}\right)\mathbf{W}_{3,z}+\mathbf{R}_{3,z}\mathbf{h}_{3,t-i-1}+\mathbf{b}_{3,z}\right)
\end{equation}
thus, $\mathbf{z}_{1,t-i}$ is the output layer of a neural network
that receives, as inputs, the external data $r_{t-i}$, multiplied
by the synaptic weights $\mathbf{W}_{1,z}\in\mathbb{R}^{d_{1}}$ ,
a recurrent internal connection $\mathbf{h}_{1,t-i-1}$, with associated
synaptic weights $\mathbf{R}_{1,z}$, plus a bias $\mathbf{b}_{1,z}$.
For the DPU module $\mathbf{z}_{3,t-i}$, we have the same structure
except that the input data is the squared deviation $\hat{\Pi}_{i}\mathbf{s}_{t}^{2}=\left(r_{t-i}-\mu_{t-i}\right)^{2}$.

Considering the whole set of equations for the memory cells update,
for each of the two modules, the corresponding LSTM has a sequential
processing that begins by outputing new candidate memory contents
for the LSTM cells $\mathbf{z}_{k,t-i}$, as well as for the input
and forget gates, in this way, the new memory contents are adaptively
processed, by the memory cells' layer so that the cells incorporate
the new memory contents to a greater or smaller degree, depending
upon the input and forget gates, that are, themselves, output layers
of respective neural networks.

For the MPU, this means that the memory cells are able to adapt to
the order and values of the elements in the sequence contained in
the time window vector $\mathbf{r}_{t}$, rather than computing the
external data as a single block, the values contained in $\mathbf{r}_{t}$
and the sequence in which they occur are incorporated in the neural
processing, which means that the network is able to adapt to changes
in patterns, including seasonalities or complex nonlinearities, that
may introduce (nonlinear) recurrence structures in data as well as
nonlinear long memory.

Likewise, the DPU memory cells are able to adapt to the order and
values of the past squared deviations in the time window vector $\mathbf{s}_{t}^{2}$,
which is particularly important for dealing with processes with nonstationary
variance that exhibit a past dependence such as nonlinear dynamical
heteroskedasticity and long memory in volatility, which are characteristic
of financial markets \cite{key-18,key-19}.

In the update of the memory cells' states, the LSTM recurrent network
is using three modules: a neural network, for processing the new memory
contents $\mathbf{z}_{k,t-i}$, and the neural networks for the input
$\mathbf{i}_{k,t-i}$ and forget $\mathbf{f}_{k,t-i}$ gates, which
lead to an adaptive management of the degree to which each memory
cell incorporates the new memory contents.

Once the memory cells' new state is processed, the LSTM network proceeds
to update the memory response modulator neurons $\mathbf{o}_{k,t-i}$,
which are output neurons of another neural network (another module).
Finally, the hidden layer is updated by the pointwise product between
the memory response modulator neurons' state and the memory cells'
state.

After processing each element in the sequence, the last state of the
LSTM network's hidden layer is fed forward to a final neuron which
updates its state in accordance with the scheme introduced previously,
namely, in this case, applying equation (20), we get, for the expected
value and variance at time $t+1$, the following relations: 
\begin{equation}
\mu_{t+1}=\tanh\left\{ \mathbf{W}_{1}^{T}\left[\mathbf{o}_{1,t}\odot\tanh\left(\mathbf{c}_{1,t}\right)\right]+\mathbf{b}_{1}\right\} 
\end{equation}
\begin{equation}
\sigma_{t+1}^{2}=\max\left\{ \left(\mathbf{W}_{3}^{T}\left[\mathbf{o}_{3,t}\odot\tanh\left(\mathbf{c}_{3,t}\right)\right]+\mathbf{b}_{3}\right),0\right\} 
\end{equation}
where, as defined before, we used the hyperbolic tangent for the next
period's returns expected value, and the ReLU activation for the next
period's expected squared deviation (the next period's variance).

Addressing the modularity of the LSTM architecture, within the context
of an MNL system, is key in both illustrating the development of scaling
artificial neural modular structures, as well as distinguishing between
two types of modularity, characterized in terms of their respective
functionality, a single target modularity and a multiple linked targets'
modularity.

In an MNL system, with learning modules comprised of LSTM learning
units, the two types of modularities are linked. Indeed, in the above
example, both the MPU and the DPU are modular recurrent networks that
are aimed each at predicting a specific target, they are each specialized
learning units, so that the modules that comprise their respective
recurrent neural networks are functionally connected towards each
of these learning units' specific objective, that is, predicting these
learning units' respective targets.

However, while the MPU is a sensory unit, in the MNL system, with
a target that does not depend on any other modular unit, the DPU depends
upon the MPU to solve its specific task. In this case, the MPU feeds
forward its outputs (its predictions) to a sliding window memory block
in the MNL system, which keeps the record of past squared deviations
updated and always matching the time window of values, in this way,
the past prediction performance of the MPU, measured in terms of the
squared deviations, can be used by the DPU to predict its target,
but this means that the DPU is working with the outputs of another
module in the system to solve its specific task.

There is a scaling functionality in a general MNL system, which is
illustrated here, where a module (in this case, the DPU) works with
a function of the outputs of another module (the MPU) to solve a different
task from the MPU. On the other hand, the DPU, like the MPU, has an
internal modular structure which is linked to its LSTM architecture.

The LSTM architecture of the DPU, as shown in the above equations,
is actually working with the past squared deviations between $r_{t-i}$
and $\mu_{t-i}$ as inputs. This shows that the entire neural processing
of the MPU is incorporated at each step of the LSTM updates of the
DPU, introducing a hierarchical modularity where the DPU depends upon
the MPU, which is characteristic of an MNL system.

On the other hand, since the LSTM architecture is itself a modular
network trained on a specific target, we have a scaling modularity
down to the individual modules, where each learning unit is, itself,
a modular network, with the difference that, while at the level of
the MNL system, each module is specialized in a specific task, and
trained to solve that task, at the level of each learning unit, under
the LSTM architecture, the (sub)modules are connected in a network
that is trying to solve a single task and the (sub)modules are being
trained conjointly towards solving that single task, which makes the
LSTM learning unit a specialized unit in this MNL system, comprised
of (sub)modules that work towards a single specialized goal.

The effectiveness of a general MNL system depends upon the training
of each module, in the sense that a module that is at a level higher
in the modular hierarchy must learn to adapt to other modules that
are already adapted/fit to solve their respective (specialized) tasks.
As stated before, this means that the learning units must be trained
following the hierarchy of connections between the subproblems, so
that, in this case, the DPU is trained after the MPU is trained.

In the present case, there is an additional specific relation between
the two modules, which is linked to the target of the DPU. Since the
DPU is aimed at predicting market volatility, in terms of a forward
looking expected squared deviation, not only does it work with outputs
produced by the MPU, but its target is also dependent on those outputs,
since it is learning to predict the next period's dispersion around
the MPU's next period prediction, this is specific of the current
example and not a general feature of MNL systems.

\subsubsection{Gated Recurrent Unit Architecture}

The GRU is an alternative to the LSTM, working without memory cells
\cite{key-41,key-42}. The hidden state update rule is such that the
GRU has an incorporated (adaptive) recurrent dynamics that modulates
the flow of information inside the unit, such that the state update
for the hidden layer is a linear interpolation between the previous
hidden state activation plus a new candidate activation update. Formally,
for the MPU and DPU, we have the following general rule:
\begin{equation}
\mathbf{h}_{k,t-i}=\left(\mathbf{1}_{d_{k}}-\mathbf{u}_{k,t-i}\right)\odot\mathbf{h}_{k,t-i-1}+\mathbf{u}_{k,t-i}\odot\tilde{\mathbf{h}}_{k,t-i}
\end{equation}
where the update gates, for the MPU and DPU, are, respectively, defined
as follows:
\begin{equation}
\mathbf{u}_{1,t-i}=\textrm{sigm}\left(\left(\hat{\Pi}_{i}\mathbf{r}_{t}\right)\mathbf{W}_{1,u}+\mathbf{R}_{1,u}\mathbf{h}_{1,t-i-1}\right)
\end{equation}
\begin{equation}
\mathbf{u}_{3,t-i}=\textrm{sigm}\left(\left(\hat{\Pi}_{i}\mathbf{s}_{t}^{2}\right)\mathbf{W}_{3,u}+\mathbf{R}_{3,u}\mathbf{h}_{3,t-i-1}\right)
\end{equation}

On the other hand, the new candidate activations are also the outputs
of neural networks which, following the GRU architecture \cite{key-42},
for the MPU and DPU, are defined, respectively, as:
\begin{equation}
\tilde{\mathbf{h}}_{1,t-i}=\tanh\left(\left(\hat{\Pi}_{i}\mathbf{r}_{t}\right)\mathbf{W}_{1,h}+\mathbf{R}_{1,h}\left(\mathbf{e}_{1,t-i}\odot\mathbf{h}_{1,t-i-1}\right)\right)
\end{equation}

\begin{equation}
\tilde{\mathbf{h}}_{3,t-i}=\tanh\left(\left(\hat{\Pi}_{i}\mathbf{s}_{t}^{2}\right)\mathbf{W}_{3,h}+\mathbf{R}_{3,h}\left(\mathbf{e}_{3,t-i}\odot\mathbf{h}_{3,t-i-1}\right)\right)
\end{equation}
where $\mathbf{e}_{1,t-i}$ and $\mathbf{e}_{3,t-i}$ are reset gates
defined as:
\begin{equation}
\mathbf{e}_{1,t-i}=\textrm{sigm}\left(\left(\hat{\Pi}_{i}\mathbf{r}_{t}\right)\mathbf{W}_{1,r}+\mathbf{R}_{1,r}\mathbf{h}_{1,t-i-1}\right)
\end{equation}
\begin{equation}
\mathbf{e}_{3,t-i}=\textrm{sigm}\left(\left(\hat{\Pi}_{i}\mathbf{s}_{t}^{2}\right)\mathbf{W}_{3,r}+\mathbf{R}_{3,r}\mathbf{h}_{3,t-i-1}\right)
\end{equation}

Considering the above equations, both for the MPU and DPU, the corresponding
GRU is comprised of three modules, the first two regard the computation
of the candidate activation $\tilde{\mathbf{h}}_{k,t-i}$ and the
update gate $\mathbf{u}_{k,t-i}$, the third is the reset gate $\mathbf{e}_{k,t-i}$.

Each of the three modules that determine the GRU's hidden layer update,
is a neural network, with output neurons, respectively, $\tilde{\mathbf{h}}_{k,t-i}$,
$\mathbf{u}_{k,t-i}$ and $\mathbf{e}_{k,t-i}$. Again, and as in
the LSTM case, the specific connectivity introduced by the MNL system,
leading to a modular (neural) hierarchy, becomes explicit, namely,
for each of the three modules that comprise the GRUs, integraded,
respectively, in the MPU and the DPU neural networks, there is a computation
of a signal external to the respective unit.

In the case of the MPU, since it is a sensory module, this signal
is given by the sequence of values $\left\{ \Pi_{i}\mathbf{r}_{t}=r_{t-i}\right\} _{i=\tau}^{0}$,
which corresponds to a past sequence observations of the financial
returns $r_{t}$, over a $\tau+1$ time window.

In the case of the DPU, on the other hand, this signal comes from
the sliding window memory module that computes the squared deviations
between the past observations $\mathbf{r}_{t}$ and the corresponding
MPU predictions, over the same temporal window, so that, for the DPU,
the signal is given by the sequence of values $\left\{ \Pi_{i}\mathbf{s}_{t}^{2}=\left(r_{t-i}-\mu_{t-i}\right)^{2}\right\} _{i=\tau}^{0}$.

Thus, for each module that comprises the DPU's GRU, we effectively
have a processing of a past time window of squared deviations around
the predictions from the MPU, stored in the sliding window working
memory module. At each level of the GRU processing, in the case of
the DPU, the MNL system's connectivity introduces an internal processing
based upon the outputs of another learning unit of the MNL system,
which also has a corresponding GRU architecture.

As in the LSTM architecture, each GRU is a modular system of neural
networks optimized to predict a specific target, while the entire
MNL system can be addressed as a modular system of modular neural
networks.

Considering the GRU modules, we have a content updating rule with
less modules than the LSTM. This is because the GRU does not have
memory cells, rather, to compute the new hidden layer's state $\mathbf{h}_{k,t-i}$,
the network uses an adaptive linear interpolation between the previous
hidden layer's state $\mathbf{h}_{k,t-i-1}$ and a candidate hidden
layer's new state $\tilde{\mathbf{h}}_{k,t-i}$, exposing the whole
state each time \cite{key-42} and such that the candidate hidden
layer's state $\tilde{\mathbf{h}}_{k,t-i}$ is computed using the
new signal in the sequence ($\hat{\Pi}_{i}\mathbf{r}_{t}=r_{t-i}$,
in the case of the MPU, and $\hat{\Pi}_{i}\mathbf{s}_{t}^{2}=\left(r_{t-i}-\mu_{t-i}\right)^{2}$,
in the case of the DPU), with synaptic weights $\mathbf{W}_{k,h}$,
plus a signal that comes from a connection with the hidden layer,
therefore, processing the previous hidden layer's state $\mathbf{h}_{k,t-i-1}$
modulated by a reset gate $\mathbf{e}_{k,t-i}$. If the reset gate
is equal to a vector of zeroes, then, in the case of the MPU, the
candidate hidden state neural network just feeds forward the new signal
to the candidate hidden layer, with associated synaptic weights $\mathbf{W}_{k,h}\in\mathbb{R}^{d_{1}}$,
applying a pointwise $\tanh$ activation to update the candidate hidden
state.

In general, the reset gate neurons adaptively modulate the exposure
of the candidate hidden state neural network to the previous hidden
state, with associated neural weights $\mathbf{R}_{k,h}\in\mathbb{R}^{d_{k}\times d_{k}}$.
The adaptive modulation results from the fact that the reset gate
is updated at each sequence step, in accordance with the output of
a respective neural network that, in the case of the MPU, processes,
as input, the sequence element $\hat{\Pi}_{i}\mathbf{r}_{t}=r_{t-i}$,
with associated neural weights $\mathbf{W}_{1,r}\in\mathbb{R}^{d_{1}}$,
and the previous hidden state $\mathbf{h}_{1,t-i-1}$, with associated
neural weights $\mathbf{R}_{1,r}\in\mathbb{R}^{d_{1}\times d_{1}}$,
applying, for the reset state state update, the pointwise sigmoid
activation. Likewise, for the DPU, it processes the sequence element
$\hat{\Pi}_{i}\mathbf{s}_{t}^{2}=\left(r_{t-i}-\mu_{t-i}\right)^{2}$,
with associated neural weights $\mathbf{W}_{3,r}\in\mathbb{R}^{d_{3}}$,
and the previous hidden state $\mathbf{h}_{3,t-i-1}$, with associated
neural weights $\mathbf{R}_{3,r}\in\mathbb{R}^{d_{3}\times d_{3}}$,
applying, for the reset state update, the pointwise sigmoid activation.

Finally, regarding the update gate, if $\mathbf{u}_{k,t-i}$ is close
to a vector of zeroes, the hidden state remains unchanged with respect
to its previous value, on the other hand, when $\mathbf{u}_{k,t-i}$
is close to a vector of ones, the hidden state update is dominated
by the new candidate activation, as per equation (34).

The update gate is, in turn, the output of a neural network, such
that, for the MPU, the pointwise sigmoid activation is applied to
an input signal where the sequence element $\hat{\Pi}_{i}\mathbf{r}_{t}=r_{t-i}$
is processed, with associated neural weights $\mathbf{W}_{1,u}\in\mathbb{R}^{d_{1}}$,
and the previous hidden state $\mathbf{h}_{1,t-i-1}$ is processed,
with associated neural weights $\mathbf{R}_{1,u}\in\mathbb{R}^{d_{1}\times d_{1}}$.
Likewise, for the DPU, the corresponding neural network is such that
the pointwise sigmoid activation is applied to an input signal where
the sequence element $\hat{\Pi}_{i}\mathbf{s}_{t}^{2}=\left(r_{t-i}-\mu_{t-i}\right)^{2}$
is processed, with associated neural weights $\mathbf{W}_{3,u}\in\mathbb{R}^{d_{3}}$,
and the previous hidden state $\mathbf{h}_{3,t-i-1}$ is processed,
with associated neural weights $\mathbf{R}_{3,u}\in\mathbb{R}^{d_{3}\times d_{3}}$.

As in the LSTM case, after processing each element in the sequence,
the last state of the GRU network's hidden layer is fedforward to
a final neuron which updates its state in accordance with the scheme
introduced previously, namely, applying equation (34), we get, for
the expected value and variance at time $t+1$, the following relations:
\begin{equation}
\mu_{t+1}=\tanh\left\{ \mathbf{W}_{1}^{T}\left[\left(\mathbf{1}_{d_{1}}-\mathbf{u}_{1,t}\right)\odot\mathbf{h}_{1,t-1}+\mathbf{u}_{1,t}\odot\tilde{\mathbf{h}}_{1,t}\right]+\mathbf{b}_{1}\right\} 
\end{equation}
\begin{equation}
\sigma_{t+1}^{2}=\max\left\{ \mathbf{W}_{3}^{T}\left[\left(\mathbf{1}_{d_{3}}-\mathbf{u}_{3,t}\right)\odot\mathbf{h}_{3,t-1}+\mathbf{u}_{3,t}\odot\tilde{\mathbf{h}}_{3,t}\right]+\mathbf{b}_{3},0\right\} 
\end{equation}
which can be contrasted with those of the LSTM in equations (32) and
(33).

Having introduced the two main architectures, we will now address
the empirical results from applying these two architectures to different
financial data.

\section{Empirical Results}

We now apply the risk and returns' predictor agent to the following
financial series:
\begin{itemize}
\item Dividend adjusted closing value for the S\&P 500 from January, 3,
1950 to January, 4, 2018, leading to 17112 data points in the logarithmic
returns series, with the first 8000 used for training, and the remaining
data points used for testing;
\item Dividend adjusted closing value for the NASDAQ Composite from February,
5, 1971 to January, 4, 2018, leading to 11834 data points in the logarithmic
returns series, with the first 5000 data points used for training,
and the remaining data points used for testing;
\item Dividend adjusted closing value for the Russell 2000 from September,
10, 1987 to January 4, 2018, leading to 7642 data points in the logarithmic
returns series, with the first 3500 data points used for training,
and the remaining data points used for testing; 
\end{itemize}
The first two series are long time series and a benchmark for EMH
proponents. A main argument for the EMH would be that the S\&P 500
is based on the market capitalization of 500 large companies and that
it is free of possible trends in price that might be used by technical
analysts. The NASDAQ Composite is also a reference, in this case,
mostly linked to technology-based companies.

The third series is the Russell 2000. While the S\&P 500 index is
a benchmark for large capitalization stocks, the Russell 2000 is a
benchmark for small capitalization stocks (\emph{small caps}), used
mainly by\emph{ small cap} funds. Although it is a smaller time series
than the previous two indices, it is still a large dataset.

For training, we need to take into account the performance of both
modules. In order to do so, we set up a randomly initialized agent
population, train each agent and select the agent with a highest score
with the score defined in terms of prediction accuracy.

Therefore, for each architecture, we build a population of randomly
initialized agents, with glorot uniform initialization. We calculate
for each agent's MPU and DPU the Root Mean-Squared Error (RMSE) and
the respective prediction accuracy defined, for the MPU, as:
\begin{equation}
p_{\varepsilon,MPU}=\frac{\#\left\{ \left|\mu_{t}-r_{t}\right|<\varepsilon:t=1,2,...,T\right\} }{T}
\end{equation}
while, for the DPU, we have:
\begin{equation}
p_{\varepsilon^{2},DPU}=\frac{\#\left\{ \left|\sigma_{t}^{2}-s_{t}^{2}\right|<\varepsilon^{2}:t=1,2,...,T\right\} }{T}
\end{equation}
where $T$ is the size of the training dataset. 

Thus, the prediction accuracy is defined as the proportion of times
in which the prediction falls within an open neighborhood of the target
variable, with size $\varepsilon$, in the case of the MPU, and $\varepsilon^{2}$,
in the case of the DPU. 

We set the DPU radius to $\varepsilon^{2}$, in order to have comparable
results between the MPU and DPU respective scales. Namely, the DPU's
target is in squared units of the MPU's target, which means that the
comparable neighborhood needs to be rescaled accordingly, so that
if we took the square root, then, we would get the same neighborhood
size, with respect to the returns' target. The two accuracies are
combined in the score which takes the mean accuracy:
\begin{equation}
score=\frac{p_{\varepsilon,MPU}+p_{\varepsilon^{2},DPU}}{2}
\end{equation}
the agent with the highest prediction score in training is chosen
for testing.

For optimization we use RMSProp and sequential minibatch training.
The RMSPRop is used with a squared error loss function, L2 norm regularization
and gradient clipping with truncation. Sequential minibatch is used
since we are dealing with a time series, which means that we need
to use a non-overlapping sequence of minibatches for training the
modules, in order to preserve the sequential nature of the data and
capture possible nonlinear patterns in past returns.

The hidden layer's width used is 2 in both architectures, this size
was set because in repeated experiments we found that the turbulent
nature of the data, and the fact that the neural network is fed sequential
1D signals, leads to a network performance such that an increase in
the hidden layer width makes the agent fit better the laminar periods,
while losing predictability with respect to the turbulent bursts,
so that the overall performance measures tended to drop on all the
series, for a hidden layer width larger than two. Therefore, we report
the results obtained for the hidden layer width equal to 2.

The main goal for the artificial agent, as defined previously, is
to produce an adaptive interval prediction for the logarithmic returns,
incorporating the volatility risk, therefore, in order to specifically
test the EMH, we begin by using a small prediction window, namely
we use the last two trading sessions and choose the best performers
of each architecture in the training data, using the score defined
above, evaluating the success of these agents in interval prediction,
measuring, for the training and test data, the proportion of cases
that fall within the interval bounds $p_{l}$ (as per equation (19)).

Then, in order to evaluate the effects of the temporal window size,
and get reproducible results, we store the highest scoring agent's
initial pre-training parameters and just change the temporal window
size, analyzing the effects of that increase on $p_{l}$ for both
the training and test data.

This process allows us to evaluate a fundamental point with respect
to financial theory: if the EMH is right, then the MNL system cannot
perform well on the short-term window (the two trading days), but
even if short-term deviations to efficiency were to occur, as we increase
the temporal window, the agent should perform increasingly worse,
since long-term returns' windows cannot have relevant information
to the next period's returns. Therefore, the artifical agent would
be adjusting to unrelated noise and including increasingly unrelated
noise in its training, such a drop in performance should be visible,
in particular, with respect to the main purpose of the artificial
agent: the interval prediction of financial returns, which uses both
the expected value and the one-period ahead standard deviation, produced
by the agent's modular networked system. A drop in interval prediction
performance, with increasing time window, should, in this case, be
more evident in the test data.

If a two-day window provides good performance and the interval prediction
performance does not drop by much with the window size, then, this
is strong evidence against the EMH and in favor of the CMH and AMH
proponents, as well as those working on machine learning-based algorithmic
investing since there is a higher predictability of financial returns
from past returns than that predicted by the EMH, which means that
there may be algorithmically exploitable patterns. Furthermore, if
rise in interval prediction performance, with window size, is obtained
then this is even stronger evidence against the EMH.

Considering, then, first, the training data results, these vary for
the two architectures and datasets. In table 1, the amplitude of the
RMSE in training, for 100 randomly initialized agents is shown to
be larger in the GRU than in the LSTM architecture, for the MPU, while,
for the DPU, it is the GRU that has the lower amplitude. Although,
in regards to the minimum, we get, for each index, similar results
for the two architectures. The performance in training seems also
to decrease with the size of the time series, so that the time series
with the longest training data exhibits lower error in training than
the time series with the shortest training data.

\begin{table}[H]
\begin{raggedright}
{\small{}}%
\begin{tabular}{|c|c|c|c|}
\hline 
{\small{}LSTM} & {\small{}Min} & {\small{}Max} & {\small{}Mean}\tabularnewline
\hline 
\hline 
\multicolumn{4}{|c|}{{\small{}S\&P 500}}\tabularnewline
\hline 
{\small{}MPU} & {\small{}0.007567} & {\small{}0.052401} & {\small{}0.014251}\tabularnewline
\hline 
{\small{}DPU} & {\small{}0.000155} & {\small{}0.048826} & {\small{}0.002587}\tabularnewline
\hline 
\multicolumn{4}{|c|}{NASDAQ}\tabularnewline
\hline 
{\small{}MPU} & {\small{}0.007880} & {\small{}0.056915} & {\small{}0.015267}\tabularnewline
\hline 
{\small{}DPU} & {\small{}0.000277} & {\small{}0.101699} & {\small{}0.006076}\tabularnewline
\hline 
\multicolumn{4}{|c|}{{\small{}Russell 2000}}\tabularnewline
\hline 
{\small{}MPU} & {\small{}0.010023} & {\small{}0.067908} & {\small{}0.018810}\tabularnewline
\hline 
{\small{}DPU} & {\small{}0.000409} & {\small{}0.139479} & {\small{}0.008414}\tabularnewline
\hline 
\end{tabular}{\small{}}%
\begin{tabular}{|c|c|c|c|}
\hline 
{\small{}GRU} & {\small{}Min} & {\small{}Max} & {\small{}Mean}\tabularnewline
\hline 
\hline 
\multicolumn{4}{|c|}{{\small{}S\&P 500}}\tabularnewline
\hline 
{\small{}MPU} & {\small{}0.007608} & {\small{}0.141089} & {\small{}0.023221}\tabularnewline
\hline 
{\small{}DPU} & {\small{}0.000160} & {\small{}0.042688} & {\small{}0.003458}\tabularnewline
\hline 
\multicolumn{4}{|c|}{NASDAQ}\tabularnewline
\hline 
{\small{}MPU} & {\small{}0.007789} & {\small{}0.152901} & {\small{}0.025152}\tabularnewline
\hline 
{\small{}DPU} & {\small{}0.000282} & {\small{}0.056534} & {\small{}0.005397}\tabularnewline
\hline 
\multicolumn{4}{|c|}{{\small{}Russell 2000}}\tabularnewline
\hline 
{\small{}MPU} & {\small{}0.010051} & {\small{}0.175307} & {\small{}0.030350}\tabularnewline
\hline 
{\small{}DPU} & {\small{}0.000414} & {\small{}0.075175} & {\small{}0.007461}\tabularnewline
\hline 
\end{tabular}
\par\end{raggedright}{\small \par}
\caption{Training data RMSE, from 100 randomly initialized agents using glorot
uniform distribution, a time window of 2, and hidden neurons layer
width of 2, for LSTM and GRU architectures. The L2 norm regularization
parameter was set, in both cases, to 0.001, the minibatch size was
set to 100, the trade-off factor for current and previous gradients
was set to 0.95, the increasing factor for learning rate adjustment
was set to 1.5 and the decreasing factor was set to 0.1. The minimum
and maximum scales allowed for the initial learning rate were, respectively,
0.001 and 0.1.}
\end{table}

In regards to the prediction accuracy, for a 1\% neighborhood size,
in the case of the MPU, and 0.01\%, in the case of the DPU, as shown
in table 2, the amplitude is very large, for each index and architecture,
ranging from low prediction accuracy to high prediction accuracy.
The two architectures, however, exhibit similar results with respect
to each index.

The accuracy results already show a deviation with respect to the
EMH. In the case of the three indices, the maximum MPU accuracy in
the training data is higher than 80\%, for a 1\% neighborhood size,
which means that the MPU is capturing a pattern in the training data,
linking the next period's returns to the last two trading days' returns,
being capable of predicting the next period's returns, within a 1\%
range, with a higher than 80\% accuracy. A similar pattern holds for
the DPU accuracy, which implies that there are patterns in the market
volatility that are also being captured by the MNL system in the training
data.

\begin{table}[H]
\begin{raggedright}
{\small{}}%
\begin{tabular}{|c|c|c|c|}
\hline 
{\small{}LSTM} & {\small{}Min} & {\small{}Max} & {\small{}Mean}\tabularnewline
\hline 
\hline 
\multicolumn{4}{|c|}{{\small{}S\&P 500}}\tabularnewline
\hline 
{\small{}MPU} & {\small{}0.192673} & {\small{}0.855464} & {\small{}0.669282}\tabularnewline
\hline 
{\small{}DPU} & {\small{}0.016633} & {\small{}0.871436} & {\small{}0.567719}\tabularnewline
\hline 
\multicolumn{4}{|c|}{NASDAQ}\tabularnewline
\hline 
{\small{}MPU} & {\small{}0.181273} & {\small{}0.872549} & {\small{}0.671791}\tabularnewline
\hline 
{\small{}DPU} & {\small{}0.014211} & {\small{}0.886309} & {\small{}0.577956}\tabularnewline
\hline 
\multicolumn{4}{|c|}{{\small{}Russell 2000}}\tabularnewline
\hline 
{\small{}MPU} & {\small{}0.175243} & {\small{}0.817038} & {\small{}0.642004}\tabularnewline
\hline 
{\small{}DPU} & {\small{}0.014874} & {\small{}0.833238} & {\small{}0.544640}\tabularnewline
\hline 
\end{tabular}{\small{}}%
\begin{tabular}{|c|c|c|c|}
\hline 
{\small{}GRU} & {\small{}Min} & {\small{}Max} & {\small{}Mean}\tabularnewline
\hline 
\hline 
\multicolumn{4}{|c|}{{\small{}S\&P 500}}\tabularnewline
\hline 
{\small{}MPU} & {\small{}0.066767} & {\small{}0.857089} & {\small{}0.573361}\tabularnewline
\hline 
{\small{}DPU} & {\small{}0.066783} & {\small{}0.862181} & {\small{}0.534920}\tabularnewline
\hline 
\multicolumn{4}{|c|}{NASDAQ}\tabularnewline
\hline 
{\small{}MPU} & {\small{}0.058423} & {\small{}0.873749} & {\small{}0.573601}\tabularnewline
\hline 
{\small{}DPU} & {\small{}0.058447} & {\small{}0.878102} & {\small{}0.534037}\tabularnewline
\hline 
\multicolumn{4}{|c|}{{\small{}Russell 2000}}\tabularnewline
\hline 
{\small{}MPU} & {\small{}0.054889} & {\small{}0.818754} & {\small{}0.545563}\tabularnewline
\hline 
{\small{}DPU} & {\small{}0.054920} & {\small{}0.815503} & {\small{}0.508249}\tabularnewline
\hline 
\end{tabular}
\par\end{raggedright}{\small \par}
\caption{Accuracy measures for table 1's agents, for an open neighborhood of
1\% for the MPU and of 0.01\% for the DPU.}
\end{table}

Now, storing the pre-training parameters for the agent with highest
score (as per equation (45)) for each index and architecture, and
considering the interval prediction results, using one unit of standard
deviation (as per equation (18)), in the case of the LSTM architecture,
for a memory window of 2 trading days, the prediction success rates,
measured in terms of the proportion of cases where the returns fall
within the interval bounds, are low for each of the three indices,
in both training and testing, as shown in table 3. The values of the
training and test data are close to each other, which shows some robustness
in terms of the stability of the results.

\begin{table}[H]
{\small{}}%
\begin{tabular}{|c|c|c|c|c|c|c|}
\hline 
\multirow{2}{*}{{\small{}$\tau+1$}} & \multicolumn{2}{c|}{{\small{}S\&P 500}} & \multicolumn{2}{c|}{NASDAQ} & \multicolumn{2}{c|}{{\small{}Russell}}\tabularnewline
\cline{2-7} 
 & {\small{}Train} & {\small{}Test} & {\small{}Train} & {\small{}Test} & {\small{}Train} & {\small{}Test}\tabularnewline
\hline 
\hline 
{\small{}2} & {\small{}0.347174} & {\small{}0.361480} & {\small{}0.401521} & {\small{}0.412798} & {\small{}0.346682} & {\small{}0.346628}\tabularnewline
\hline 
{\small{}5} & {\small{}0.426658} & {\small{}0.442039} & {\small{}0.512425} & {\small{}0.520446} & {\small{}0.422923} & {\small{}0.426531}\tabularnewline
\hline 
{\small{}10} & {\small{}0.444862} & {\small{}0.457045} & {\small{}0.542369} & {\small{}0.546455} & {\small{}0.438218} & {\small{}0.441398}\tabularnewline
\hline 
{\small{}15} & {\small{}0.445922} & {\small{}0.458320} & {\small{}0.544064} & {\small{}0.550345} & {\small{}0.440058} & {\small{}0.443201}\tabularnewline
\hline 
{\small{}20} & {\small{}0.445980} & {\small{}0.458604} & {\small{}0.544355} & {\small{}0.550861} & {\small{}0.439595} & {\small{}0.443063}\tabularnewline
\hline 
{\small{}25} & {\small{}0.445660} & {\small{}0.458448} & {\small{}0.544242} & {\small{}0.550641} & {\small{}0.439420} & {\small{}0.442923}\tabularnewline
\hline 
{\small{}30} & {\small{}0.445718} & {\small{}0.458402} & {\small{}0.544737} & {\small{}0.551011} & {\small{}0.438953} & {\small{}0.442784}\tabularnewline
\hline 
\end{tabular}{\small \par}

\caption{Interval prediction success in training and test data for agent with
LSTM architecture, with increasing window size, using the same initial
pre-training parameters as the two trading sessions window highest
scoring agent.}
\end{table}

Even though each module exhibits good results, in terms of predicting
their respective targets, the interval prediction fails more, which
means that the interplay between the expected value and volatility,
for a time window of 2 trading days, is not capable of capturing the
majority of returns swings.

This might be assumed, in first sight, to be supportive of the EMH,
however, this is, in fact, related to the ability of the neural network
to capture the turbulent patterns in the market, since there is a
basic tension between two dynamics, characteristic of financial returns'
turbulence: a laminar dynamics, where the market shows small returns
movements, and turbulence bursts, where the market exhibits sudden
and clustering volatility with large price jumps.

A neural network may capture very well the smaller swings, and even
present good results on average in the prediction, but fail in capturing
the joint expected value and volatility dynamics, so that the interval
prediction fails more often than it is right. For a two period window,
in the case of the LSTM, the prediction success rates are less than
40\% (near 35\%) in training and testing for the S\&P 500 and Russell
2000 indexes, and it is slightly higher than 40\% in the case of the
NASDAQ.

Thus, while some predictability may be present, for a two-day window,
an interval prediction fails more often than not. If the time window
increases, however, we can see a tendency for the increase in the
interval prediction success rates.

Therefore, for agents with a cognitive architecure given by the LSTM
modular system, with the same initial parameters, increase in the
window size, leads, in this case, to a rise in the interval prediction
success rates, in both training and testing, which goes against the
hypothesis that long term data should have no bearing in the prediction,
and, due to increasing noise, that it should in fact lead to a drop
in prediction accuracy. The rise in prediction success, however, still
leads to poor performance in the case of the S\&P 500 and the Russel
2000, which exhibit success rates lower than 50\% and higher than
40\%. The only index for which the corresponding agent exhibits an
accuracy higher than 50\% is the NASDAQ Composite, with accuracy rates
of 54.4737\% (in training) and 55.1011\% (in testing).

These results change, in terms of values, for the GRU architecture
as shown in table 4. Indeed, while, for a window size of 2 trading
days, the GRU shows a worse performance in interval prediction than
the LSTM, that performance rises as the window size is increased,
in both training and testing, and it rises higher than the LSTM. Indeed,
for a 20 day window, the GRU architecture exhibits interval prediction
success rates higher than 60\%, in both training and testing data,
for all the three indices, and for a 30 day window, the prediction
success rate is higher than 80\% for all the three indices, in both
training and testing.

\begin{table}[H]
{\small{}}%
\begin{tabular}{|c|c|c|c|c|c|c|}
\hline 
\multirow{2}{*}{{\small{}$\tau+1$}} & \multicolumn{2}{c|}{{\small{}S\&P 500}} & \multicolumn{2}{c|}{NASDAQ} & \multicolumn{2}{c|}{{\small{}Russell}}\tabularnewline
\cline{2-7} 
 & {\small{}Train} & {\small{}Test} & {\small{}Train} & {\small{}Test} & {\small{}Train} & {\small{}Test}\tabularnewline
\hline 
\hline 
{\small{}2} & {\small{}0.310530} & {\small{}0.307346} & {\small{}0.309047} & {\small{}0.303119} & {\small{}0.313215} & {\small{}0.303602}\tabularnewline
\hline 
{\small{}5} & {\small{}0.389987} & {\small{}0.387760} & {\small{}0.381964} & {\small{}0.380771} & {\small{}0.388252} & {\small{}0.379085}\tabularnewline
\hline 
{\small{}10} & {\small{}0.494361} & {\small{}0.494775} & {\small{}0.491566} & {\small{}0.484662} & {\small{}0.495690} & {\small{}0.477797}\tabularnewline
\hline 
{\small{}15} & {\small{}0.590715} & {\small{}0.590794} & {\small{}0.589537} & {\small{}0.586947} & {\small{}0.593948} & {\small{}0.577475}\tabularnewline
\hline 
{\small{}20} & {\small{}0.675251} & {\small{}0.677875} & {\small{}0.679435} & {\small{}0.671868} & {\small{}0.685838} & {\small{}0.664472}\tabularnewline
\hline 
{\small{}25} & {\small{}0.762516} & {\small{}0.761285} & {\small{}0.766465} & {\small{}0.758219} & {\small{}0.772464} & {\small{}0.752872}\tabularnewline
\hline 
{\small{}30} & {\small{}0.830982} & {\small{}0.834825} & {\small{}0.835223} & {\small{}0.835080} & {\small{}0.840698} & {\small{}0.826268}\tabularnewline
\hline 
\end{tabular}{\small \par}

\caption{Interval prediction success in training and test data for agents with
GRU architecture, with increasing window size, using the same initial
pre-training parameters as the two trading sessions window highest
scoring agent.}
\end{table}

Both results place in question the EMH in the sense that, not only
do past returns' data have an impact in interval prediction, but,
for the case of the GRU architecture, we can predict the financial
returns, within an interval bound comprised of an expected value and
square root of the expected squared deviation, with more than 80\%
success, that is, we have an agent that is capable of using a forward
looking expected returns and volatility expectation to produce an
interval prediction that is right more than 80\% of the time.

This result was obtained setting $l$ to $1$ in the interval (as
per equation (18)). If we increase $l$, for the same 30 trading days
window, we get higher success rates. Namely, as shown in table 6,
the interval prediction, for $l=1.5$, has a success rates near 95\%
in test data for all the three indices, and when $l=2$, we get success
rates near 98\%, in test data, for all the three indices. The training
and test data success rates also show values close to each other which
is evidence of robustness of the results.

\begin{table}[H]
\begin{centering}
\begin{tabular}{|c|c|c|c|c|}
\hline 
{\small{}\#SD} & {\small{}Data} & {\small{}S\&P 500} & NASDAQ & {\small{}Russell}\tabularnewline
\hline 
\hline 
\multirow{2}{*}{{\small{}1 }} & {\small{}Train} & {\small{}0.830982} & {\small{}0.835223} & {\small{}0.840698}\tabularnewline
\cline{2-5} 
 & {\small{}Test} & {\small{}0.834825} & {\small{}0.835080} & {\small{}0.826268}\tabularnewline
\hline 
\multirow{2}{*}{{\small{}1.5}} & {\small{}Train} & {\small{}0.945340} & {\small{}0.941093} & {\small{}0.940116}\tabularnewline
\cline{2-5} 
 & {\small{}Test} & {\small{}0.945641} & {\small{}0.946995} & {\small{}0.950502}\tabularnewline
\hline 
\multirow{2}{*}{{\small{}2}} & {\small{}Train} & {\small{}0.980605} & {\small{}0.979150} & {\small{}0.976163}\tabularnewline
\cline{2-5} 
 & {\small{}Test} & {\small{}0.979892} & {\small{}0.979625} & {\small{}0.984073}\tabularnewline
\hline 
\end{tabular}
\par\end{centering}
\caption{Interval prediction success rates for training and test data for agents
with GRU architecture with different units of standard deviation,
using the same initial pre-training parameters as the two trading
sessions window highest scoring agent.}
\end{table}

From a financial theory perspective, these results have a high relevance.
In the case of the S\&P 500, the data goes from the 1950s to 2018,
this means that we have different economic and financial environments.

However the agent consistently performs with approximately 98\% performance
over these different periods. In the training data, such performance
can be expected, as long as the learning approximates well the main
features of the data. However, for the test data, in the case of the
S\&P 500, the agent was able to exhibit a higher than 90\% prediction
success rate (for the case of an interval prediction using a unit
of $1.5\sigma_{t+1}$ and $2\sigma_{t+1}$ ) in a period that goes
from 1981 to 2018. Furthermore, the agent was able to achieve this
performance, having been trained in the period that goes from 1950s
up to 1981.

This indicates the presence of some structural nature in the stock
market that remains the same and is captured by the MNL system, allowing
it to consistently perform over long time periods. Figure 1 shows
the series and prediction interval bounds for the S\&P 500 in both
training (left graph) and test data (right graph).

\begin{figure}[H]
\includegraphics[scale=0.5]{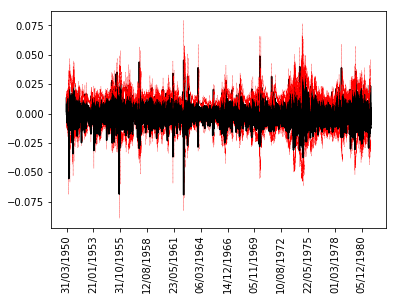}\includegraphics[scale=0.5]{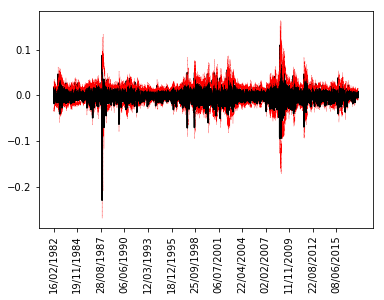}

\caption{S\&P 500 prediction interval bounds (in red) and observed returns
data (in black), for both training (left) and test data (right) series,
with $l=1.5$ and 30 days past window.}
\end{figure}

The two graphs show the main reason for the success of the agent,
namely, whenever there is a turbulence burst, with volatility clustering,
the agent's predicted interval also becomes wider, so that the volatility
bursts and price jumps tend to be within the interval's bounds. Likewise,
whenever the market enters into a low volatility period, the agent's
predictions also anticipate low volatility, and, thus, the interval
bounds tend to become smaller.

The agent has thus captured an emergent nonlinear dynamics present
in the index that links the last 30 days' returns and volatility to
the next period's returns and volatility. This dynamics must be structural,
since the agent's predictions, in the case of the S\&P 500 hold, with
a similar success measure, for different samples (training and test)
that cover different historical periods, both economic and financial,
which is consistent with the hypothesis that the nonlinear dynamics,
linked to financial turbulence, may be intrinsic to the trading patterns
and the nature of the market as a very large open network of adaptive
agents that process financial information and trade based on their
expectations for market dynamics.

Although, microscopically, different agents may trade based on different
rules, at the macro-level there is an emergent pattern in the market
that leads to a nonlinear dynamics that makes turbulence patterns
in financial returns predictable enough to be captured effectively
by an agent equipped with a modular networked system of GRUs.

Although the LSTM architecture is also effective in capturing some
pattern, the GRU approximates better the large jumps and buildup associated
to the volatility bursts, as well as the gradual reduction in volatility
associated with laminar periods. These results are consistent with
the AMH, which addressess complex nonlinear returns' dynamics as a
result of collective emergent patterns from evolutionary trading rules
and actions of financial agents.

It is important to stress that these agents are trying to anticipate
the result of their own collective trading behavior, in the sense
that they form expectations of market rise (\emph{bullish}) or market
fall (\emph{bearish}), so that the market exhibits an emergent adaptive
dynamics with complex feedback, this makes emerge complex collective
behaviors that are structurally linked to the market as an adaptive
systemic network.

Furthermore, the test data includes the period of financial automation,
which took off since 2008. So that, the agent's training, which ended
in 1981, works well even in a transition period and a post-automation
wave trading ecosystem.

The results obtained for the other two indices reinforce these findings,
namely, in the case of the NASDAQ, as shown in figure 2, the buildups
in volatility tend to be accompanied by widening of the prediction
interval, and the less volatile periods also tend to be periods with
smaller prediction interval amplitudes, so that the agent's interval
predictions exhibit an antecipatory adaptive response to market conditions,
anticipating the returns' turbulent and laminar dynamics, therefore,
the majority of returns fluctuations fit within the interval bounds.

\begin{figure}[H]
\includegraphics[scale=0.5]{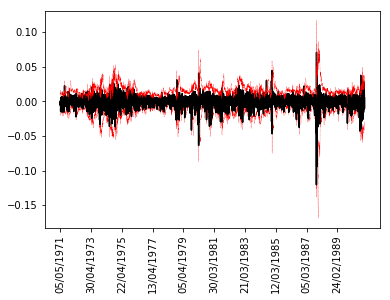}\includegraphics[scale=0.5]{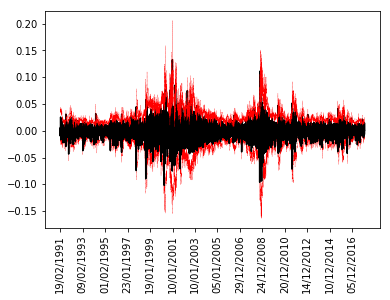}

\caption{NASDAQ Composite prediction interval bounds (in red) and observed
returns data (in black), for both training (left) and test data (right)
series, with $l=1.5$ and 30 days past window.}
\end{figure}

The NASDAQ focuses strongly on technology stocks, thus, the agent
performs well in both a benchmark for \emph{large caps}, that also
constitutes a benchmark for market efficiency, as well as in a more
sector focused benchmark. This is strong evidence against the EMH,
and in favor of complexity science-based market theories.

The last index, reinforces these results, in that we have a \emph{small
caps} index, and a smaller series, but we still get a high interval
prediction success rate, and a similar coincidence between the agent's
prediction interval amplitudes and the market volatility patterns.

\begin{figure}[H]
\includegraphics[scale=0.5]{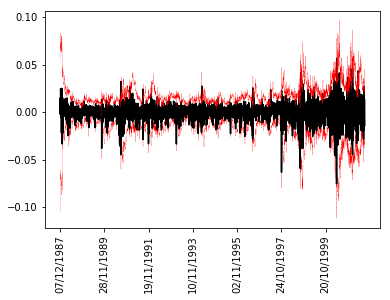}\includegraphics[scale=0.5]{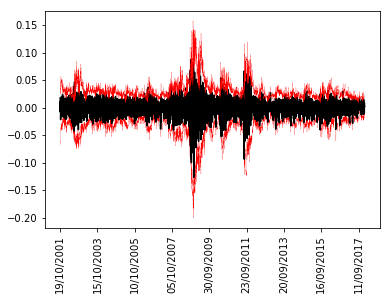}

\caption{Russell 2000 prediction interval bounds (in red) and observed returns
data (in black), for both training (left) and test data (right) series,
with $l=1.5$ and 30 days past window. }
\end{figure}

The Russel 2000 index provides an important addition to the above
results, since we might expect the agent to perform better in small
caps and sectorial indices than in the large caps reference, that
is the S\&P 500 index, but this is not the case, in all three indices,
we get similar results both in terms of interval prediction success
rate as well as in terms of turbulent and laminar periods' anticipation.

\section{Conclusion}

Machine learning allows for the development of AI systems with interconnected
learning modules, these are MNL systems that are able to solve complex
problems comprised of multiple linked subproblems, where each cognitive
module is specialized and trained on a specific task, but the modules
are linked in a network that maps the network of connections between
the different subproblems.

In the present work, we illustrated an application of AI with an MNL
cognitive system to finance. In financial markets, the market dynamics
is driven by expectations on returns on one's investment and expectations
on the risk associated with the returns on one's investment. These
are the two fundamental drivers of traders' decision making.

An MNL system is capable of integrating these two drivers in an expectation
for the next period's returns, based on a time window of past returns,
and an expectation for the next period's squared deviation around
the next period's returns expectation, based on a time window of past
squared deviations.

We introduced an artificial agent with such a cognitive system, which
learns to predict the next period's returns and the next period's
squared deviation around its next period's returns prediction and
uses the two parameters to provide an interval prediction for the
next period's returns, calculating the square root on the next period's
predicted squared deviation as a time-varying volatility parameter
that is used to build a prediction interval centered on the expected
returns. In this way, we have an interval that incorporates both the
expectation around the returns and the expectation around risk.

According to the EMH, such an artificial intelligent system should
not be successful in returns' prediction, that is, such an interval-based
prediction should not work. On the other hand, in accordance with
the CMH and AMH, this interval-based prediction should be possible,
since there are emergent collective trading patterns coming from financial
agents' adaptive trading dynamics, which lead to a nonlinear dynamics
with stochastic and deterministic components that explain the presence
of long memory in volatility and financial turbulence. 

The application of the MNL system to the S\&P 500, NASDAQ Composite
and Russel 2000 shows that, not only, does interval prediction work,
but the predictability increases with an increase in the temporal
window. Thus, for a GRU-based MNL system and a 30 trading days window
there is a high predictability of returns, which is higher than what
we get for smaller time windows, in both training and test data. Although
the LSTM does not perform as well as the GRU, we also get this effect
of increasing predictability with window size.

The prediction performance is linked to the ability of the MNL system
to capture main emergent nonlinearities and patterns in volatility,
so that the prediction interval produced by the artificial agent is
a dynamical interval that is capable of adapting to market conditions
contracting when it anticipates low volatility periods and expanding
when it anticipates high volatility periods.

Regarding financial management, these results help explain the success
of algorithmic investing, and it lends support to alternative trading
schools, in particular technical analysis, which assumes that patterns
are present in past financial time series, and that has been a key
component in recent efforts of application of machine learning, including
recurrent neural networks, to finance \cite{key-43,key-44,key-45}.

However, there is an important point to be made: the results hold
for each index, independently of the trading school, indeed, we did
not explicitly use any technical analysis elements, which would deviate
from our goal of pattern discovery, and EMH testing, since it would
introduce, top-down, a specific trading school, deviating the agent
from the basic pattern discovery which was under test.

The agent is, therefore, not addressing technical analysis specifically,
but is, instead, working with two major financial parameters used
by the investment management school that is associated with the EMH:
the expected value and the variance. This is a purposeful key feature,
since the goal was to explicitly test the EMH, especially the weak
efficiency hypothesis' statistical submartingale assumption, against
a machine learning alternative. If the weak form of the EMH fails,
the semi-strong and the strong versions of the EMH also fail.

The main results that we obtained, with an artificial agent equipped
with a GRU-based MNL cognitive system, show that there are long-term
patterns in past financial returns that allow an artificial agent
to successfully produce an interval prediction of returns, adjusting
the expected value and the volatility predictions to the market conditions,
consistent with the main arguments of the complexity approaches to
finance \cite{key-11,key-12,key-36,key-37,key-46}.

The success of this machine learning alternative means that the instruments
and techniques used by the EMH-based investment management school
need to be revised and transformed, incorporating machine learning,
and working with forward-looking expected values and variances. In
this sense, traditional financial theory may not hold with respect
to the EMH assumption but some of its techniques, especially in regards
to risk management and portfolio management, including the expected
value and variance-based portfolio optimization techniques, developed
in the 1950s by Markowitz \cite{key-47,key-48}, may still survive
to the degree to which they can be incorporated in successful machine
learning algorithms, capable of capturing the markets' main emergent
nonlinear dynamics.

\end{document}